\documentclass[lettersize,journal]{IEEEtran}

\IEEEoverridecommandlockouts                              
\usepackage{amsmath} 
\usepackage{amssymb}  
\usepackage{graphicx}
\usepackage[inkscapelatex=false]{svg}
\usepackage[numbers,sort&compress]{natbib}
\usepackage{colortbl}
\usepackage{threeparttable}

\title{\LARGE \bf
\textit{Phy-Tac}: Toward Human-Like Grasping via \textit{Phy}sics-Conditioned \textit{Tac}tile Goals
}
\author{Shipeng Lyu$^{\dag}$, Lijie Sheng$^{\dag}$, Fangyuan Wang, Wenyao Zhang, Weiwei Lin, Zhenzhong Jia$^{*}$\\
David Navarro-Alarcon$^{*}$ and Guodong Guo

\thanks{S. Lyu, F. Wang, and D. Navarro-Alarcon are with the Department of Mechanical Engineering, The Hong Kong Polytechnic University (PolyU), Kowloon, Hong Kong. D. Navarro-Alarcon is also with the Research Institute for Smart Ageing (RISA), PolyU.}%
\thanks{W. zhang, W. Lin and G. Guo are with the Ningbo Institute of Digital Twin, Eastern Institute of Technology (EIT), China.}%
\thanks{L. Sheng and Z. Jia are with the Department of Mechanical and Energy Engineering, Southern University of Science and Technology, Shenzhen, China.}%
\thanks{$\dag$ Equal contribution; * Corresponding author}
}

\begin{document}

\maketitle
\thispagestyle{empty}
\pagestyle{empty}

\begin{abstract}
Humans naturally grasp objects with minimal level required force for stability, whereas robots often rely on rigid, over-squeezing control.
To narrow this gap, we propose a human-inspired physics-conditioned tactile method (Phy-Tac) for force-optimal stable grasping (FOSG) that unifies pose selection, tactile prediction, and force regulation.
A physics-based pose selector first identifies feasible contact regions with optimal force distribution based on surface geometry. 
Then, a physics-conditioned latent diffusion model (Phy-LDM) predicts the tactile imprint under FOSG target. 
Last, a latent-space LQR controller drives the gripper toward this tactile imprint with minimal actuation, preventing unnecessary compression.
Trained on a physics-conditioned tactile dataset covering diverse objects and contact conditions, the proposed Phy-LDM achieves superior tactile prediction accuracy, while the Phy-Tac outperforms fixed-force and GraspNet-based baselines in grasp stability and force efficiency.
Experiments on classical robotic platforms demonstrate force-efficient and adaptive manipulation that bridges the gap between robotic and human grasping.
\end{abstract}

\section{Introduction}
\label{sec:Introduction}
Humans naturally achieve stable grasping through active tactile regulation that applies only the necessary amount of force to maintain stability while avoiding object damage~\cite{newbury2023deep}.
In contrast, robotic grasping still relies heavily on rigid control or over-squeezing strategies, which compromise safety, energy efficiency, and adaptability, especially when interacting with fragile or compliant materials~\cite{mao2024efficient}. 
Therefore, achieving force-optimal stable grasping (FOSG), where the grasp remains stable with the minimal necessary contact force, remains a fundamental challenge for robotic manipulation.

Most existing approaches treat grasping pose planner and force regulation as separate stages.
Pose planners like~\cite{graspnet} usually optimize geometric force-closure or heuristic confidence, without reasoning about whether the chosen contacts can sustain an optimal force distribution. 
On the other hand, force regulations often intervene only after slip occurs or just provide a fixed contact force rather than proactively regulating contact force toward an optimal regime~\cite{yang2024vt}. 
Consequently, robots lack a unified mechanism that jointly reasons over contact planning and contact force optimization within a single loop, thereby preventing globally consistent force optimization and perception-control coordination.

\begin{figure}[t]
    \centering
    \includegraphics[width=\linewidth]{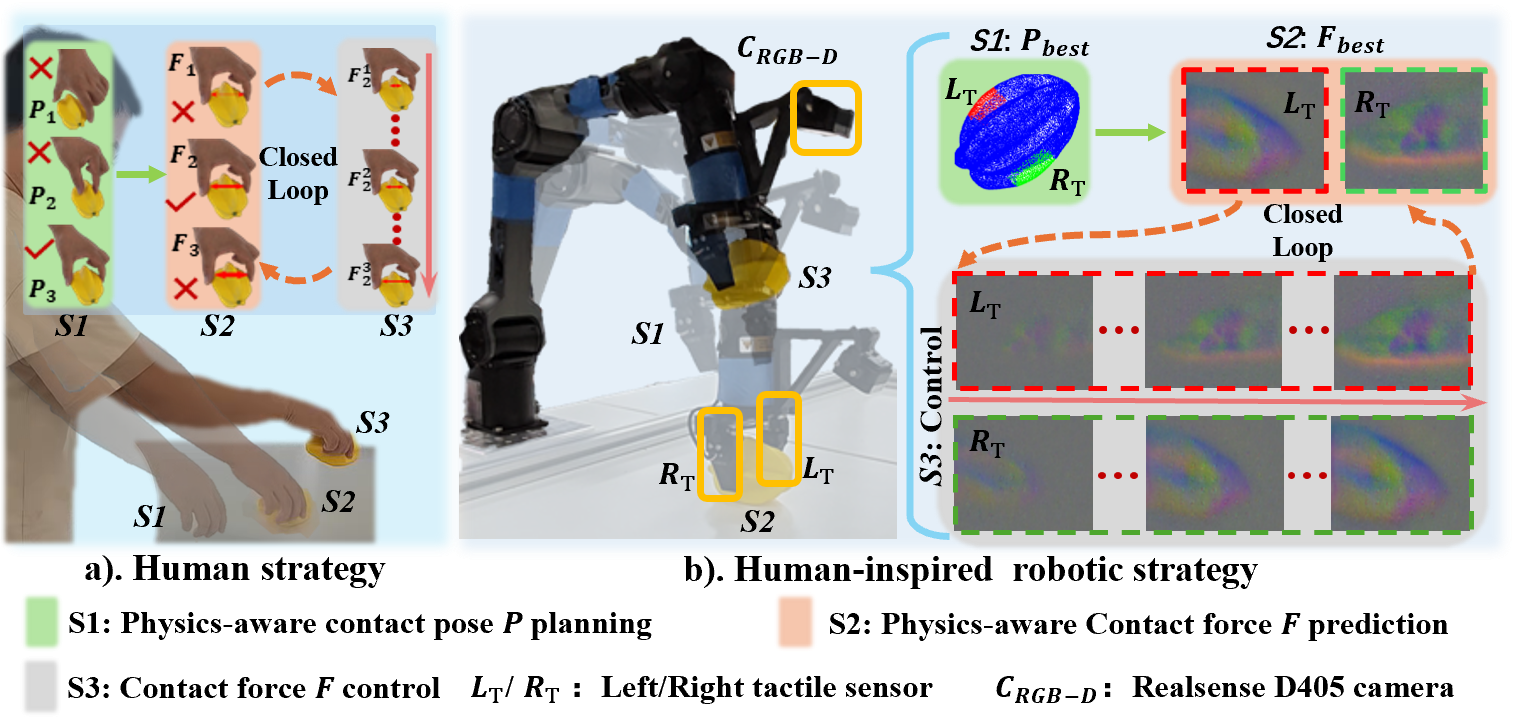}
    \caption{
    Comparison of human grasping strategy (a) and our Phy-Tac framework (b) for FOSG.
    Humans first select an optimal contact pose (\textit{S1}), estimate the required grasping force (\textit{S2}), and refine it to a just-enough level to maintain stability (\textit{S3}).
    Likewise, Phy-Tac unifies pose planning (\textit{S1}), tactile state prediction (\textit{S2}), and force regulation (\textit{S3}) to achieve the same principle, where \textit{S2} and \textit{S3} form a closed loop for optimal force regulation.
    }
    \label{fig:Descrip}
\end{figure}

However, humans achieve force-optimal grasping through an anticipatory and feedback-rich process as shown in Fig.~\ref{fig:Descrip}-a.
They first select geometrically favorable contact regions (\textit{S1}), estimate the minimal required force from prior experience (\textit{S2}), and then refine it using fingertip tactile feedback (\textit{S3}). 
Normally, humans form \textit{S2} and \textit{S2} as closed loops to directionally regulate force in time for contacting state changes to avoid sliding occurs.
Inspired by this principle, our work shown in Fig.~\ref{fig:Descrip}-b aims to endow robots with a similar capability to select grasp pose, estimate optimal force, and regulate contact force proactively rather than reactively, thereby bridging the gap between human dexterity and robotic manipulation.

\begin{figure*}[ht]
    \centering
    \includegraphics[width=\linewidth]{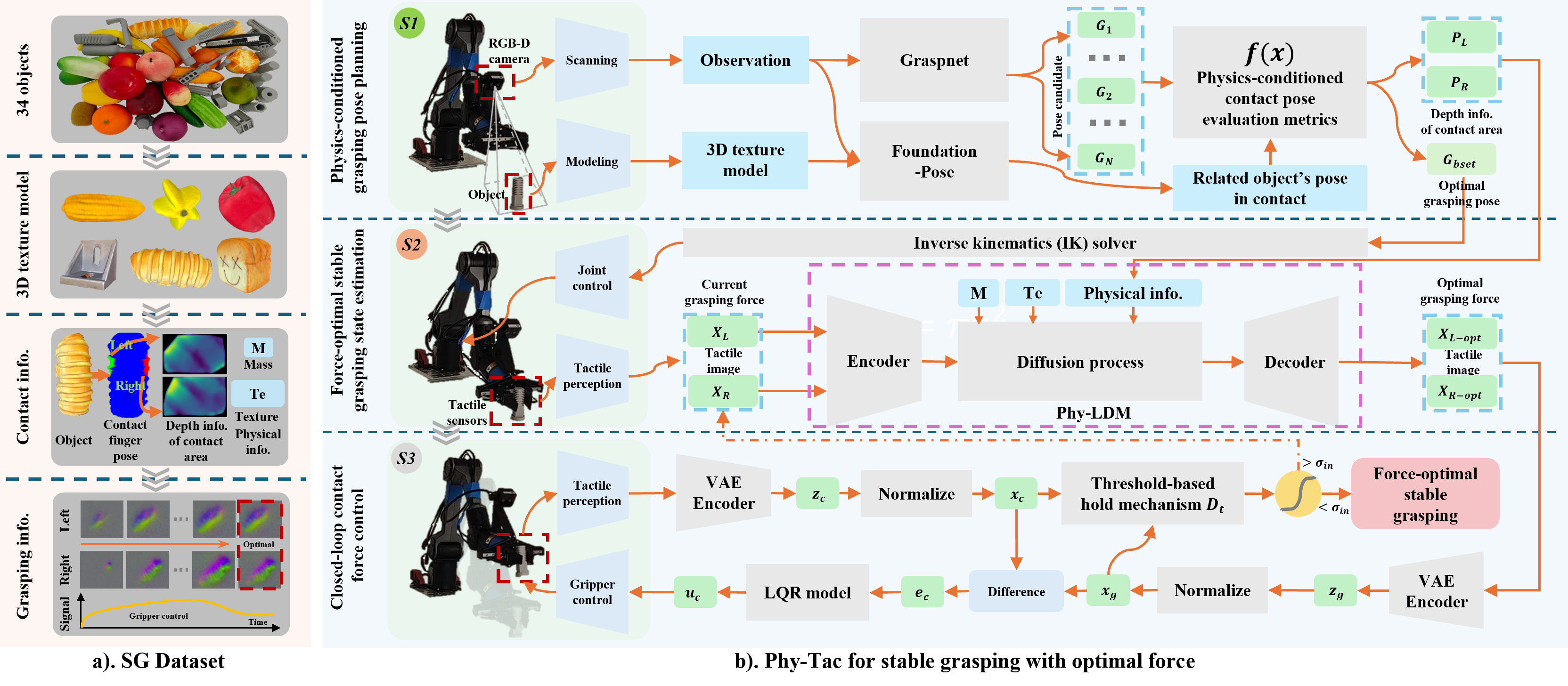}
    \caption{The description of human-inspired stable grasping method Phy-Tac. 
    a). Our physics-conditioned SG dataset consists of objects' $3$D texture models, contact region information, control signals, and tactile states.
    b). Our force-optimal stable grasping strategy Phy-Tac contains 3 steps, i.e., grasping pose planning, grasping state estimation, and contact force regulation. }
    \label{fig:Method}
\end{figure*}

To this end, we propose a human-inspired Phy-Tac for FOSG that integrates grasping pose planning, tactile prediction, and grasping force regulation within a unified tactile–control pipeline as shown in Fig.~\ref{fig:Method}. 
The key idea is to predict the tactile imprint corresponding to a FOSG and to drive the gripper toward this tactile imprint with minimal actuation. 
Specifically, a physics-conditioned contact selector identifies grasp poses with optimal force distribution, a physics-conditioned latent diffusion model (Phy-LDM) predicts the optimal tactile state, and a latent-space LQR controller efficiently converges to that state.
Our main contributions are summarized as follows:
\begin{itemize}
    \item We propose a human-like Phy-Tac,  which unifies grasp planning, tactile estimation, and force regulation into a single closed loop centered on a force-optimal tactile goal, enabling proactive and minimal-force grasping rather than reactive stabilization.
    \item We develop a physics-conditioned Phy-LDM that predicts the tactile imprint corresponding to optimal stable force, and align it through a LQR servo to achieve FOSG.
    \item We provide a physics-conditioned tactile dataset that covers objects’ physical attributes and tactile features, providing a foundation for physically consistent tactile study.
\end{itemize}

\section{Related work}
\label{sec:Related_work}

\textbf{Grasp pose planning}.
Recent end-to-end methods \cite{graspnet,wang2022dexgraspnet,weng2024dexdiffuser} leverage large-scale datasets to train grasping policy satisfying geometric force closure or maximizes pose confidence.
Unfortunately, they do not optimize contact forces with respect to the physical properties of the contact region.
Therefore, \cite{Phygrasp} proposes a physics-informed, large multimodal model that incorporates object physical information into grasp pose generation, which achieves over a $10\%$ improvement.
However, it requires the creation of extensive, task-specific datasets for training, which is both labor-intensive and costly.
Therefore, we explore a more efficient strategy to incorporate objects' geometric features into grasp pose planning.

\begin{figure}[h]
    \centering
    \includegraphics[width=\linewidth]{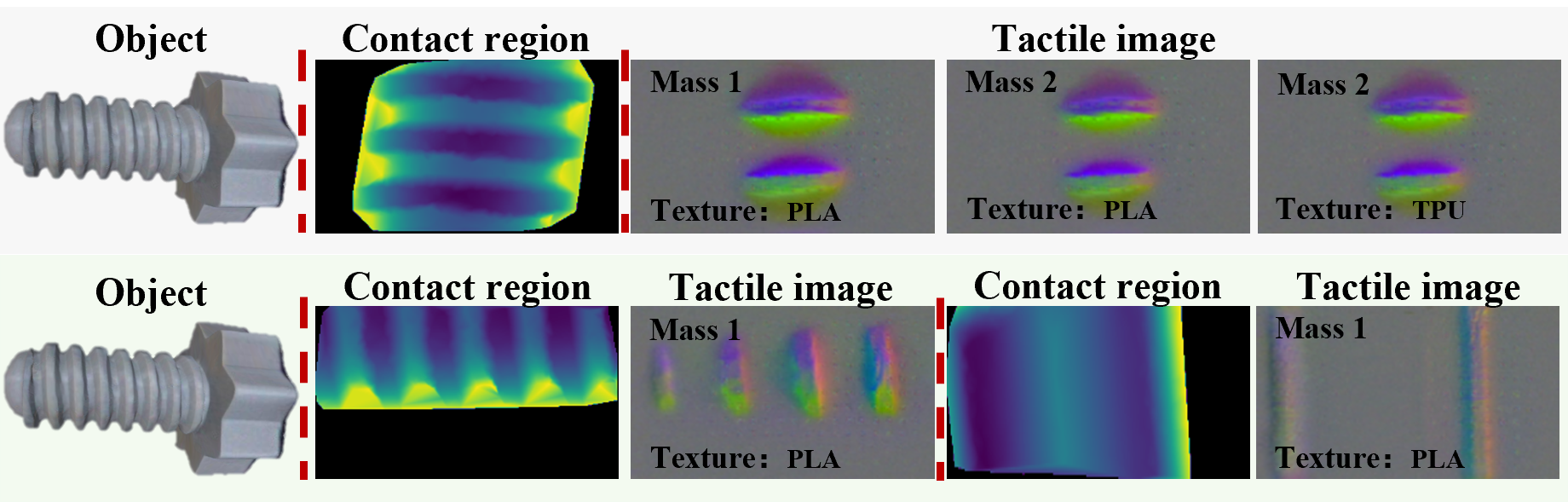}
    \caption{A example to show the influence of physical condition for contact tactile state. Specifically, the physical factors are mass $M$, texture $T$, and contact region $P_t$.}
    \label{fig:factors}
\end{figure}

\textbf{Tactile sensor in grasping}.
Tactile sensors play a critical role in force regulation during robotic grasping by providing rich contact information, e.g., force distribution.
The vision-based sensors (e.g., GelSight~\cite{yuan2017gelsight}) are particularly notable in tactile sensors~\cite{li2024comprehensive} for their unique capability to convert contact-induced deformations into high-resolution images as shown in Fig.~\ref{fig:factors}.
This property is widely used in grasping tasks, such as contact detection~\cite{liu2023enhancing, zhang2023robust, zhang2023tirgel}, and force estimation \cite{ma2019dense, sundaralingam2019robust, lin20239dtact}, and grasping control \cite{matak2022planning, costanzo2021control, lloyd2021goal}.
Motivated by these advantages, we employs a vision-based tactile sensor for FOSG task.
Although previous studies have explored context-conditioned contact estimation~\cite{tu2025texttoucher, yang2024binding, gungor2025towards}, these coarse-grained text conditions fail to estimate fine-grained trends in contact property, e.g., optimal force required for stable grasping. 
Therefore, we leverages fine-grained physical information in contact region together with the initial contact state to guide fine-grained contact force prediction.

\textbf{Diffusion model}.
Diffusion models synthesize of high-fidelity and diverse samples by progressively adding noise to data and learning the reverse denoising process~\cite{DDPM}. 
Compared with GAN-based approaches~\cite{CGAN}, diffusion models offer superior training stability, detail preservation, and controllability, which has led to their increasing adoption in robotic perception tasks.
For tactile domain, diffusion models have been applied to tactile image generation for robotic manipulation~\cite{lin2024vision, rodriguez2024touch2touch, yang2023generating}. 
These works demonstrate the strong potential of diffusion models in modeling fine-grained tactile generation task. 
Therefore, we use physics-conditioned latent diffusion model to generate the tactile imprint of the optimal stable state, providing a clear and high-fidelity tactile prior for achieving FOSG.

\textbf{Tactile dataset}.
There are many tactile datasets~\cite{yang2022touch,li2019connecting,fu2024touch} available for different usage, such as  tactile-driven image stylization.
Although these well-collected datasets have been used for grasp stability \cite{kanitkar2022poseit}, they still face several challenges.
First, they lack the detailed information of grasping objects, e.g., material and mass, which is important effector for tactile state as shown in Fig.~\ref{fig:factors}.
Second, the object $3$D information providing by third-view cameras cannot accurately describe the contact region, which reduces the prediction accuracy for optimal contact state.
Moreover, only initial and stable grasping states cannot describe the complex grasping process~\cite{nakahara2025learning}.
Therefore, we provide a physical-conditioned tactile dataset to solve these problem for FOSG study.

\section{Methodology}
\label{sec:Methodology}
We propose a human-inspired unified framework Phy-Tac for FOSG, which embeds force optimality as a guiding principle throughout the entire grasping process (Fig.~\ref{fig:Method}).
Instead of treating tactile feedback as a passive signal, our method formulates an explicit tactile goal representing the force-optimal stable state and actively drives the system toward it.
The pipeline consists of four components: Sec.~\ref{sec:PTD} builds a physics-conditioned tactile dataset; Sec.~\ref{sec:planning} plans grasp poses with contact-patch physics; Sec.~\ref{sec:OSGSE} predicts the optimal tactile imprint as the control target; and Sec.~\ref{sec:LQR} realizes that target via a latent-space servo that achieves stability with just-enough force.

\subsection{Physics-conditioned tactile dataset}
\label{sec:PTD}
To enable modeling of tactile mechanisms that respect real-world contact physics, we construct a physics-conditioned tactile dataset that explicitly couples tactile signals with object physical attributes and stable grasp states.
This dataset contains $34$ objects with per-object material label $T$, mass $M$, and an accurate $3$D model $O$ scanned by an EinScan-HX device as shown in Fig.~\ref{fig:Method}-a.
For each grasp, we extract the contact-region point cloud $P_c$ and align it to the fingertip frame $P_t$ by a rigid transform $T^C_T$.
We keep points within a rectangular fingertip window, as Eq.~\ref{Eq:contact_area}:
\begin{equation}
    P_t = \{p_t(x,y,z)|p_t =T^C_Tp_c, |x|\leq \frac{w}{2},|y|\leq \frac{h}{2} \} .
    \label{Eq:contact_area}
\end{equation}
Where, $h$ and $w$ are the hight and wight of tactile sensor contact region.
This operation removes out-of-gripper points and standardizes the contact geometry, which mainly contributes to the the tactile feature.
During contact, we synchronously record gripper commands and gripper feedback state $\{u_c,u_f\}$ together with the tactile image $x_c$  of the gripper fingertip.
All tactile signals are collected using an Xsense G1-WS device, capturing $400\times700$ RGB image at $40$Hz.
Specifically, we explicitly provide the tactile image $x_s$ in FOSG condition.
The FOSG condition is obtained by manual detection during data collection while gradually reducing the contact force until the target slips.
Generally speaking, $x_c$ is a dynamic tactile image reflecting the contact information during the grasping process (shown in Fig.~\ref{fig:Descrip}-b \textit{S3}), which is not necessarily the optimized $x_s$ (shown in Fig.~\ref{fig:Descrip}-b \textit{S2}).

Totally, our physics-conditioned tactile dataset provides about $8.6$K paired entries 
$\{O,M,T,P_t,u_c,u_r,x_c,x_s\}$ across $0.3$K stable grasps.

\subsection{Contact-optimized pose planning}
\label{sec:planning}
Human grasping behavior often involves subtle finger adjustments before applying significant gripping force to achieve uniform contact pressure, thereby reducing local stress concentration and potential slip.
Inspired by this observation, we further refine the top $N$ grasp candidates with score $S$ generated by GraspNet through introducing a human-inspired pose planning strategy thereby promoting an optimal contact force distribution.
This enables the gripper to select poses that inherently favor stable and distribution-optimal interactive force before tactile optimization begins.
Prior studies on contact modeling~\cite{Contact_G} indicate that the geometry of the contact region critically influences the distribution of grasping forces.
Specifically, sharp or highly curved regions cause localized stress concentrations and potential slippage, whereas gradual surfaces promote uniform force distribution and stable contact.
To capture these physically meaningful geometric properties, we define a physics-inspired geometric consistency metric $\mathcal{F}(\mathcal{P}_t)$ to evaluate each candidate pose $P_t$ as Eq.~\ref{Eq:Point_cost_F}.
\begin{equation}
\left\{
    \begin{array}{lr}
             \mathcal{F}(\mathcal{P}_t) = \alpha\cdot S_{rough}+\beta\cdot(1-C_N)+\gamma\cdot U_C, &  \\
             W_p = \delta \cdot (1-S) + (1-\delta)\cdot \mathcal{F}(\mathcal{P}_t). &  
    \end{array}
\right.
\label{Eq:Point_cost_F}
\end{equation}
where $S$ is the grasp candidate's score estimated by the GraspNet;
$S_{rough}$ quantifies the local surface roughness, representing the frictional stability of the contact region; 
$C_N$ measures the consistency of surface normals across contact points, indicating the alignment of potential contact forces; 
and $U_C$ describes the curvature uniformity, which reflects the evenness of stress distribution.
Together, these three descriptors approximate the mechanical plausibility of contact stability without requiring explicit stress computation.

To balance the physically motivated geometric consistency with the data-driven confidence from GraspNet, we introduce a combined heuristic $W_p$.
where $\delta$ controls the trade-off between learning-based confidence and physics-informed geometry.
A smaller $W_p$ indicates a pose that simultaneously exhibits high network confidence and strong force-distribution consistency.

\subsection{Optimal tactile imprint estimation}
\label{sec:OSGSE}
We propose a physics-conditioned latent diffusion model  (Phy-LDM) for generating tactile image for FOSG by following procedures.
\\
\textbf{Problem statement}.
Given the current tactile observation $x_c$, the depth image of the contact patch $P_t$, and the object's physical attributes (i.e., mass $M$ and texture $Te$), we aim to synthesize the tactile imprint $x_s$ of the force-optimal stable state.
Specifically, $x_s$ is the tactile image recorded in FOSG state (Set.~\ref{sec:PTD}).
We formulate state estimation as a tactile state synthesis problem by a learning generator $f_\theta$ as Eq.~\ref{Eq:problem}.
\begin{equation}
    \hat{x}_g=f_\theta(x_c,P_t,M,Te)\approx x_s .
    \label{Eq:problem}
\end{equation}
Therefore, the controller in Sec.~\ref{sec:LQR} can servo the gripper toward the corresponding goal latent with minimal actuation.
\\
\textbf{Variational latent space}.
The VAE serves as a fundamental component for learning compressed latent space of tactile images, enabling efficient downstream diffusion-based generation. 
The pair $(E,D)$ denotes the encoder–decoder of a stable-diffusion-style AutoencoderKL~\cite{LDM}. 
\begin{equation}
    \mathcal{L}_{\text{KL}}  = D_{\mathrm{KL}}\!\left(q_{\phi}(z \mid x)\,\|\,\mathcal{N}(0,I)\right).
    \label{Eq:KL}
\end{equation}
\begin{equation}
    \omega_{\text{KL}}(epoch)    = \min\!\left(1, \frac{epoch}{E_{\text{warm}}}\right)\,\lambda_{\text{KL}}.
    \label{Eq:Learning_cost}
\end{equation}
For a tactile image $x$, we obtain the latent space $z$ through $z=E(x)\in\mathcal{R}^{C_z\times H_z \times W_z}$ and the estimated tactile image by $\hat{x}=D(z)$.
The latent channel $C_z$ and spatial size are set by configuration and the down-sampling factor.
The training protocol implements a optimization strategy to balance pixel-wise reconstruction fidelity and latent regularization.
Specifically, the pixel-wise reconstruction fidelity is ensured by the L1 loss, while the latent regularization is guaranteed by the loss $\mathcal{L}_{KL}$ as shown in Eq.~\ref{Eq:KL}.
The composite loss function combines two components through linearly warmed-up KL weight $\omega_{KL}$ shown in Eq.~\ref{Eq:Learning_cost}.
Where $E_{warm}$ is the warm-up length, and $\lambda_{KL}$ is the target KL weight.
\\
\textbf{Physics-conditioned latent diffusion model}.
To explicitly couple tactile generation with grasping physics, we design a physics-conditioned Latent Diffusion Model (Phy-LDM).
Unlike conventional conditional diffusion that relies purely on visual or text cues, Phy-LDM injects physics-informed conditions, i.e., contact geometry and object attributes, into denoising step of the latent diffusion process.
This design allows the model to generate tactile goal states that obey force–balance and contacting constraints, producing stable yet minimally loaded contact patterns.
Specifically, the current tactile  and depth images are encoded as $z_t=E(x_t)$ and $z_{cdp} = E(P_t)$.
The object mass $M$ and texture $Te$ are projected into embeddings $e_M$ and $e_T$, then fused as the physical conditioning vector $C = [e_M,e_T]$.
The $C$ is injected into the U-Net via cross-attention to modulate feature propagation.

The diffusion process learns to generate optimal tactile imprint through iterative denoising, conditioned on physical constraints.
The forward diffusion process in Eq.~\ref{Eq:forward} progressively corrupts the target latent representation $z^0$ through additive noise following a variance-preserving schedule.
\begin{equation}
     z^t = \sqrt{\bar{\alpha}^t} z^0 + \sqrt{1 - \bar{\alpha}^t} \varepsilon, \quad \varepsilon \sim \mathcal{N}(0, I).
     \label{Eq:forward}
\end{equation}
The reverse process employs a DDIM sampler \cite{DDIM} to generate samples through deterministic reverse diffusion, as formalized in Eq. \ref{Eq:back}. 
\begin{equation}
    z^{t-1} = \sqrt{\frac{\bar{\alpha}^{t-1}}{\bar{\alpha}^t}} \left( z^t - \sqrt{1 - \bar{\alpha}^t} \varepsilon_\theta \right) + \sqrt{1 - \bar{\alpha}^{t-1}} \varepsilon_\theta.
    \label{Eq:back}
\end{equation}
The noise prediction $\epsilon_\theta$ in Eq.~\ref{Eq:diffusion_process} is performed by a conditioned U-Net architecture, which incorporates multi-modal conditioning through cross-attention mechanisms. 
This architecture integrates both the current grasp state $z_a$ and physical prompt information.
\begin{equation}
    \left\{
        \begin{array}{lr}
                 \epsilon_\theta = \mathrm{U\!Net}([ z_t , z_{in} , z_{cdp} ],t,C), &  \\
                 \mathcal{L}_{\text{obj}} = \mathbb{E}_{z_0, z_{in},z_{cdp},C,t, \varepsilon} \left[ \| \varepsilon - \varepsilon_\theta\|^2 \right].& 
        \end{array}
    \right.
    \label{Eq:diffusion_process}
\end{equation}
The model parameters are optimized by minimizing the objective function $\mathcal{L}_{\text{obj}}$, which ensures accurate noise prediction conditioned on the physical constraints.

\subsection{LQR Grasp Servo}
\label{sec:LQR}
To drive the robotic gripper towards a stable grasp, we formulate a closed-loop controller in the latent tactile space. The control objective is to minimize the discrepancy between the current tactile observation and the synthesized target tactile imprint, while avoiding excessive gripper motions. \\
\textbf{State representation}.
The current tactile image $x_c$ is encoded into a low-dimensional latent vector $z_c=E_p(x_c)$ by pre-trained control-aware VAE encoder.
The target tactile state $z_g$ is obtained by encoding the synthesized target tactile image $x_s$ from the Phy-LDM. 
The current latent error is defined as $e_c$.
Since the latent space is explicitly designed to capture controllable directions, the normalized control state is directly expressed as Eq. \ref{Eq:xt}. Where $S_{scale}$ denotes a scale vector obtained from the empirical distribution of latent values for normalization.
\begin{equation}
    e_c = diag(1/S_{scale})(z_c -z_g) \in R^m.
    \label{Eq:xt}
\end{equation}
\\
\textbf{LQR modeling}.
Around the goal state, the dynamics of $e_c$ under incremental gripper displacement $\Delta u_c=u_c-u_{c-1}$ are approximated by a linear model in Eq. \ref{Eq:LQR_model}
\begin{equation}
    \left\{
    \begin{array}{lr}
             e_{c+1} = Ae_c+B\Delta u_c+d+w_c, &  \\
             J=\sum_{t=0}^\infty (e^T_cQe_c+\Delta u_c^TR\Delta u_c).& 
    \end{array}
\right.
\label{Eq:LQR_model}
\end{equation}
Where $A$ and $B$ are identified from data using recursive least squares, $d$ denotes a small constant bias, and $w_t$ is process noise. 
The identified dynamics include a small constant bias $d$ and process noise $w_c$. 
In practice, we omit $d$ and $w_c$ during controller synthesis, since their effects are consistently small relative to the latent error magnitude. 
Residual bias and noise are tolerated by the robustness of the LQR feedback and by the threshold-based stopping rule introduced below.
The control objective $J$ minimizes both the latent error and the control error with the $Q\ge 0$ and $R >0$.
Solving the discrete algebraic Riccati equation, we get the control signal $\Delta u=-Ke_c$.
\\
\textbf{Threshold-based hold mechanism}.
As exact tactile matching is unattainable, a threshold-based stopping rule is introduced. The normalized error distance $D_c=\lVert e_c\lVert_2$ is monitored within a sliding window of duration $\tau$.
A grasp is considered achieved if $D_c \le \delta_{in}$ for all frames in the window and safety conditions are satisfied, and the controller switches to hold mode.
\section{Experimental study}
\label{sec:Experiment}
To verify the efficiency of our Phy-Tac method, we evaluate each procedure in our strategy by comparing with state-of--the-art methods in both qualitative and quantitative experiments.

\subsection{Physics-conditioned grasping pose planning}
\begin{figure}[h]
    \centering
    \includegraphics[width=\linewidth]{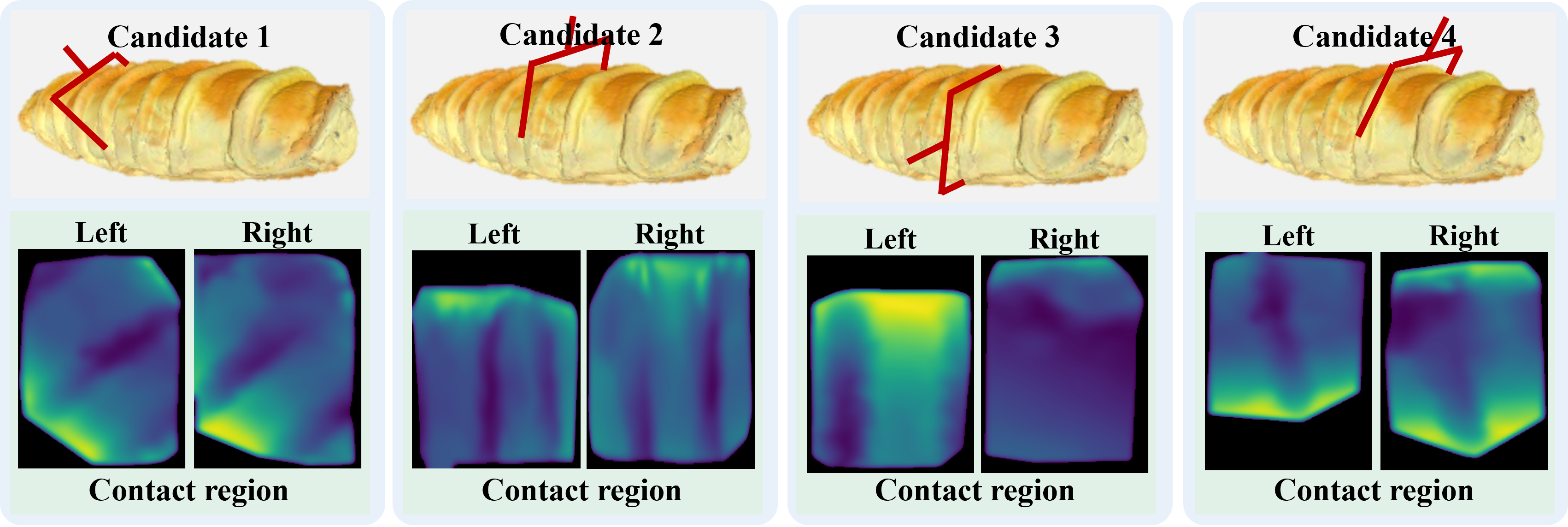}
    \caption{The contact description of the selected candidates with top score. The first row in each sub-figure is the grasping pose generated by GraspNet for bread.
    The second row is the depth information of contact region in left/right finger for generated grasping pose.}
    \label{fig:region}
\end{figure}

\begin{figure}[h]
    \centering
    \includegraphics[width=\linewidth]{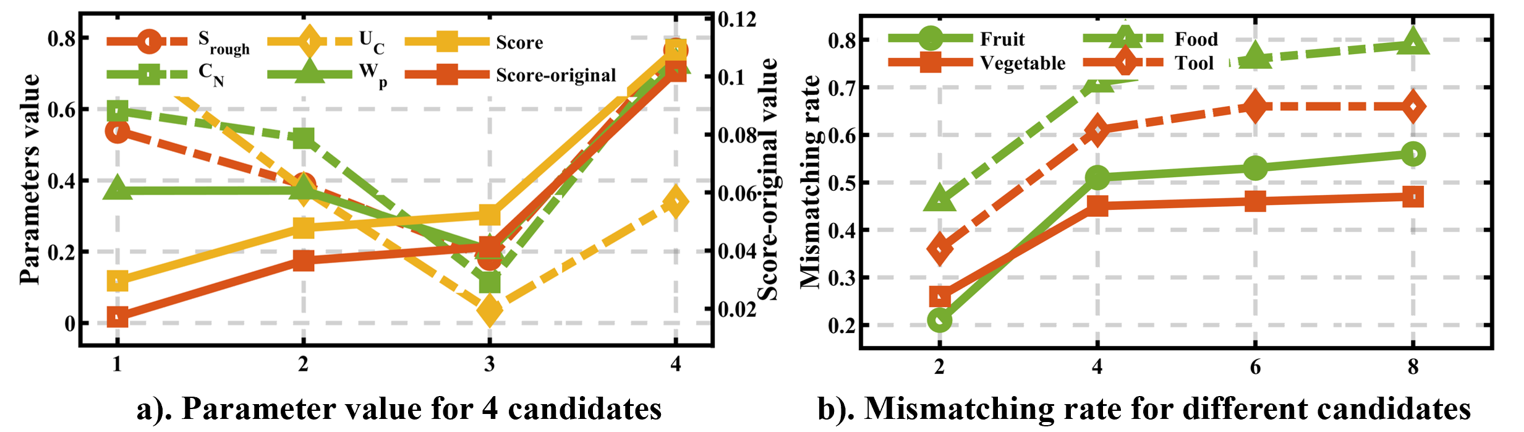}
    \caption{The parameter value of the selected grasping candidates generated by GraspNet. 
    a). The score and evaluating metrics of each candidates for grasping pose planning.
    b). The value of mismatching rate is used to select suitable candidate numbers in contact-optimal pose planning process.
    }
    \label{fig:para}
\end{figure}

As the method described in \ref{sec:planning}, we select $4$ candidates with top scores generated by GraspNet as shown in Fig.~\ref{fig:region}.
Subsequently, the corresponding depth images under contact region for each gripper finger are captured.
Visually, the contact image of first candidate is sharper and exhibits a smaller contact region when compared with the third candidate.
Therefore, this grasping attempt is more likely to slip or rotate, resulting failure of grasping based on the studies for considering contact geometry in stable grasp, e.g. \cite{Contact_G}.
To avoid this problem, a physics-conditioned function is used for selecting force distribution optimal pose for stable grasping instead of intuitive analysis.
The quantitative results in Fig.~\ref{fig:para}-a shows the detailed parameter values and the weighted cost $W_p$ for each candidate with the constant values $\alpha =0.2$, $\beta =0.6$,$\gamma  =0.2$, and $\delta = 0.5$.
Specifically, physics-conditioned parameters $S_{rough}$, $C_N$, and $U_C$ are normalized across the candidates.
Following our principle for grasping pose selection, the third candidate is the best one ($W_p\le0.2$) which is optimal for contact force distribution.

Furthermore, a significant mismatch rate ($\ge20\%$) exists between the candidate with the optimal contact force distribution (best $W_p$) and the top-ranked candidate from the grasp pose generator (best $S$) during our experiment for four object groups as shown in Fig.~\ref{fig:para}-b.
This significantly high rate ($\ge20\%$) underscores the necessity of a physics-based evaluation to identify grasps with optimal force distribution, rather than directly adopting the initial output from GraspNet.
Furthermore, the trend observed in the mismatch rate as the number of candidates increases substantiates the rationale for selecting the top four candidates from GraspNet in our experiments.
Specifically, beyond four candidates, the rate of mismatch plateaus ( $\le10\%$), indicating diminished returns with additional candidates.

\subsection{Physics-conditioned contact state generation}
\begin{figure*}[t]
    \centering
    \includegraphics[width=\linewidth]{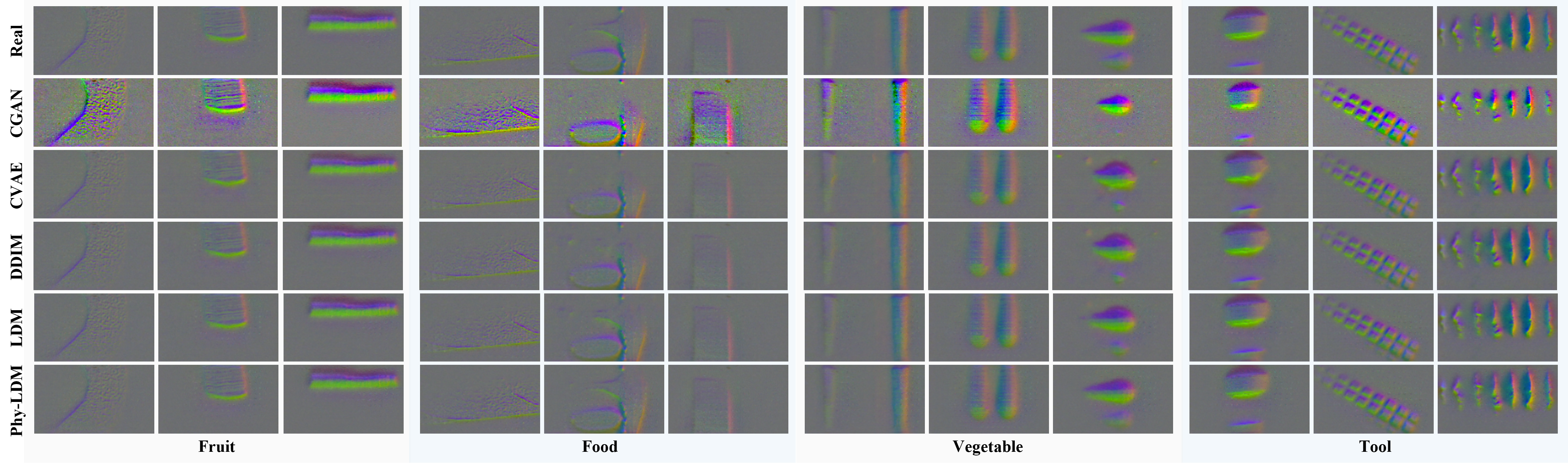}
    \caption{The predicted tactile results of different state-of-the-art methods for FOSG. This comparative results contains four type of objects, i.e., fruit, food, vegetable, and tool. }
    \label{fig:performance}
\end{figure*}

To evaluate the effectiveness of our physics-conditioned contact state (tactile image) generation method for stable grasping with optimal contact force, we compare it with several state-of-the-art generative models.
Specifically, five quantitative metrics are employed for evaluation, including pixel-level fidelity, i.e., mean absolute error (MAE) and root mean square error (RMSE), structural similarity, i.e., structural similarity index (SSIM), distribution consistency, i.e., learned perceptual image patch similarity (LPIPS), reconstruction quality, i.e., peak signal-to-noise ratio (PSNR).
\\
\textbf{Comparative study}.
Compared with the four classical image generation baselines, our Phy-LDM achieves significant and consistent improvements across all evaluation metrics, as shown in Table \ref{Table:Comp_results}.
Specifically, Phy-LDM reduces both MAE and RMSE by over $30–40\%$ compared with the best diffusion-based baseline (LDM~\cite{LDM}), and simultaneously achieves higher PSNR ($\ge42$) and SSIM ($\ge0.98$), indicating superior pixel-level accuracy and structural preservation. 
The LPIPS score also drops notably by over $17\%$, demonstrating enhanced perceptual quality and realism.
This performance gain stems from the explicit integration of contact geometry and object properties into the generative process. By embedding these physical priors into the latent space, Phy-LDM can model the intrinsic relationship between contact mechanics and tactile appearance, rather than relying solely on data-driven visual correlations. As a result, it generates tactile images that more faithfully reproduce fine-grained pressure distributions, deformation patterns, and texture variations across different contact scenarios as shown in Fig.~\ref{fig:performance}.

\begin{table}[h]
\centering
    \begin{threeparttable} 
        \caption{The comparative results with several classical image generation methods.}
        \begin{tabular}{ccccccc}
        \hline
        \rowcolor[HTML]{EFEFEF} 
        Method & MAE$\downarrow$ & RMSE$\downarrow$ & PSNR$\uparrow$ & SSIM$\uparrow$ & LPIPS$\downarrow$\\ \hline
        CGAN~\cite{CGAN} & 0.083 & 0.093 & 32.137 & 0.915  & 0.143\\
        CVAE~\cite{CVAE}   & 0.063 & 0.079 & 34.082 & 0.935  & 0.119\\
        DDIM~\cite{DDIM}   & 0.011 & 0.021 & 38.089 & 0.965 &0.085 \\
        LDM~\cite{LDM}   & 0.015 & 0.016 & 39.776 & 0.968  & 0.081\\
        \rowcolor[HTML]{EFEFEF}
        Phy-LDM     & \textbf{0.009} & \textbf{0.011} & \textbf{42.087} & \textbf{0.988}  & \textbf{0.067}\\ \hline
        \end{tabular}
        \label{Table:Comp_results}
        \begin{tablenotes} 
        \tiny
		\item Here, the LDM uses a lightweight stable-diffusion–style architecture due to limited training data. 
     \end{tablenotes} 
    \end{threeparttable} 
\end{table}

\begin{figure}
    \centering
    \includegraphics[width=\linewidth]{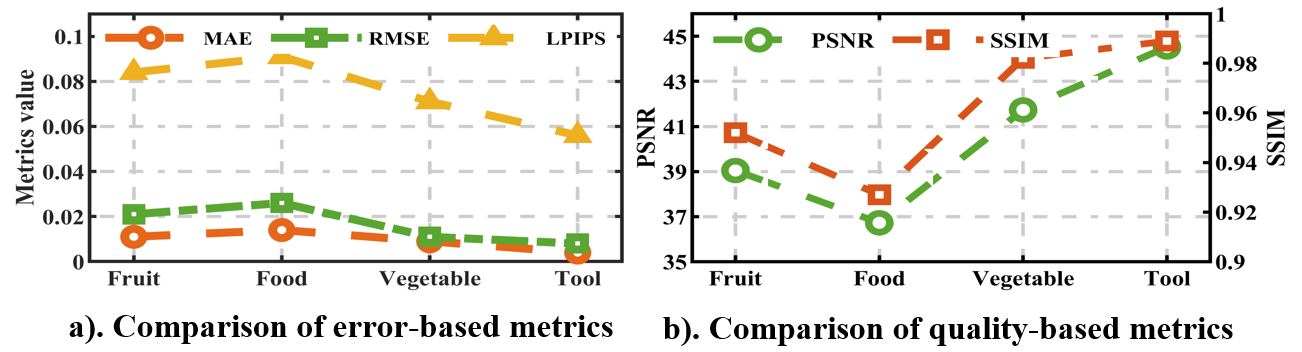}
    \caption{The evaluation value of different metrics for four types of grasp objects under our Phy-LDM method.}
    \label{fig:Metrics_values}
\end{figure}

To further validate the generalization capability of Phy-LDM across different object categories, we evaluate its performance on four representative groups: fruit, food, vegetable, and tool, as shown in Fig.~\ref{fig:Metrics_values}. 
The model consistently maintains low MAE ($\le 0.01$), RMSE ($\le 0.03$), and LPIPS ($\le 0.1$), while achieving high PSNR ($\ge 36$) and SSIM ($\ge 0.93$), demonstrating stable tactile image generation under diverse contact conditions.
Notably, Phy-LDM achieves its best performance on tool-type objects, while showing relatively lower accuracy on food objects. 
This discrepancy primarily arises from the highly deformable and soft nature of food items (e.g., bread), where the contact geometry dynamically changes during interaction. 
Although our model introduces geometric priors to constrain the generation process, the non-rigid deformation of soft materials leads to slight mismatches between predicted and real contact states. In contrast, Tool-type objects exhibit rigid and stable contact geometries, allowing Phy-LDM to leverage the encoded physical priors more effectively, resulting in sharper and more faithful tactile reconstructions.
These results highlight that the incorporation of contact geometry and material properties enables Phy-LDM to generalize well across categories, while the residual performance gap reveals potential directions for future improvements in modeling deformable object interactions.

\begin{figure}[h]
    \centering
    \includegraphics[width=\linewidth]{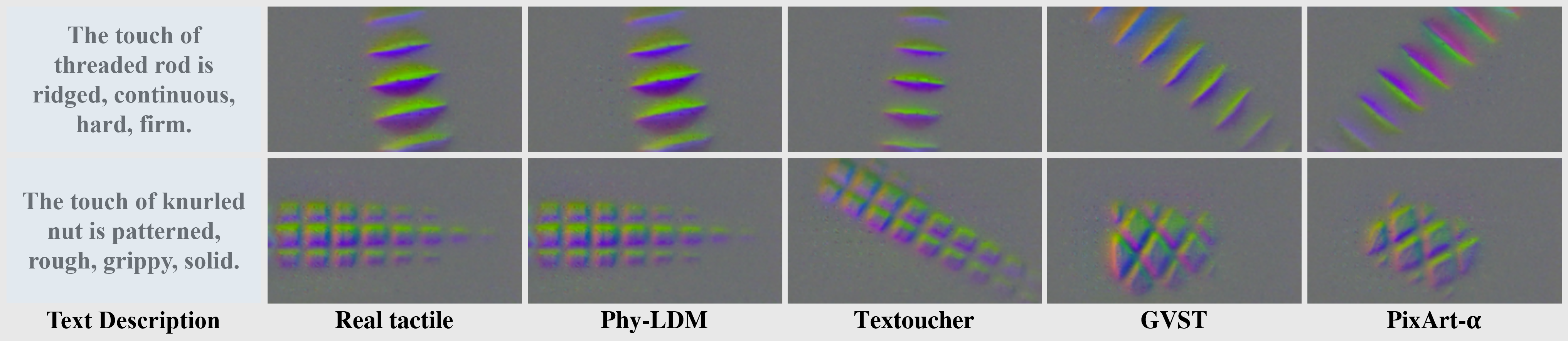}
    \caption{The comparative result between Phy-LDM and other representative methods. 
    Phy-LDM can generate more accurate tactile image based on the physical contact information.}
    \label{fig:generation2}
\end{figure}

For comparing with state-of-the-art tactile state estimation method, several experimental study is implemented.
As shown in Table~\ref{Table:Comp_results2}, our physics-conditioned method substantially outperforms existing approaches, generating more accurate tactile images for FOSG.
Specifically, Phy-LDM reduces both MAE and RMSE over $60\%$, LPIPS over $40\%$, improves the PSNR over $11\%$, SSIM over $2\%$.
Therefore, this superiority is reflected not only in pixel-level, but also in structural similarity and distribution consistency in the generated tactile image.
To visually show the each method's performance, a significant example is provided in Fig.~\ref{fig:generation2}.
Although the other method can generate tactile image closely aligning with the provided text descriptions,
they are difficult to produce one as the real tractile image in given contact situation as Phy-LDM.
The reason is that the pure text description cannot reflect the contact information in detail, especially the geometry in contact region which is one of the main factors to influence the contact pattern in tactile image.
Instead, our method introduces the fine-grained information in generation process to overcome this limitation.
Therefore, the tactile image generated by Pyh-LDM can accurately reflect the contact state. 
\begin{table}[h]
\centering
\caption{The comparative results with state-of-the-art tactile image estimation methods.}
\begin{tabular}{ccccccc}
\hline
\rowcolor[HTML]{EFEFEF} 
Method      & MAE $\downarrow$ & RMSE $\downarrow$ & PSNR $\uparrow$  & SSIM $\uparrow$  & \multicolumn{1}{c}{\cellcolor[HTML]{EFEFEF}LPIPS $\downarrow$} \\ \hline
GVST~\cite{yang2023generating}        &0.042     & 0.057     & 30.094& 0.943 & 0.235                                              \\
PixArt-$\alpha$~\cite{chen2024pixart}    & 0.037    & 0.042     & 34.438 & 0.924 & 0.197                                              \\
TextToucher & 0.031    &  0.035    & 36.814 & 0.952 & 0.142                                              \\
\rowcolor[HTML]{EFEFEF} 
Phy-LDM     & \textbf{0.010} & \textbf{0.014} & \textbf{41.087} & \textbf{0.974}  & \textbf{0.083}
\\ \hline
\end{tabular}
\label{Table:Comp_results2}
\end{table}
\\
\textbf{Ablation study}.
To analyze the contribution of each conditional component to generation performance, we conducted an ablation study as summarized in Table~\ref{Table:Ablation_results}.
The results clearly show that removing the contact geometry descriptor ($z_{cdp}$) and contact state representation ($z_{in}$) leads to the most significant performance degradation across all metrics, i,e., MAE, RMSE, and LPIPS increased by $35\%$, $31\%$, and $28\%$ respectively; PSNR and SSIM decreased by $14\%$, and $4\%$ respectively.
This observation highlights that accurate tactile generation relies heavily on spatial and structural cues describing the contact configuration, which determine the overall texture layout and fine-grained deformation patterns in tactile imprint.
In contrast, excluding the object’s physical properties, such as mass ($e_M$) and texture embedding ($e_T$), results in a smaller performance drop, i,e., MAE, RMSE, and LPIPS increased by $18\%$, $15\%$, and $12\%$ respectively; PSNR and SSIM decreased by $2\%$, and $1\%$ respectively. 
This suggests that, within the limited variability of our dataset, object-level physical attributes play a secondary but complementary role. 
They refine the realism and local consistency of generated images, but the tactile texture formation is primarily driven by geometric contact priors.
Overall, this experiment confirms that incorporating contact geometry and instantaneous interaction states is crucial for achieving physically consistent tactile image synthesis, while the integration of material and mass information further enhances perceptual fidelity and generalization.

\begin{table}[h]
\centering
\caption{The ablation results of Phy-LDM method.}
\begin{tabular}{cccccc}
\hline
\rowcolor[HTML]{EFEFEF} 
Condition & MAE$\downarrow$ & RMSE$\downarrow$ & PSNR$\uparrow$ & SSIM$\uparrow$ & LPIPS$\downarrow$ \\ \hline
Lacking $z_{cdp}$         &  0.018   &  0.019    &   34.742   &  0.942    &  0.108     \\
Lacking $z_{in}$        & 0.014    & 0.016     &  36.371     & 0.955     &  0.094      \\
Lacking $e_T$      & 0.012    & 0.013     &  39.932    & 0.969     &  0.079     \\
Lacking $e_M$      &  0.011   &   0.013   & 40.882     &  0.976    &  0.077     \\
\rowcolor[HTML]{EFEFEF}
Phy-LDM     & \textbf{0.009} & \textbf{0.011} & \textbf{42.087} & \textbf{0.988}  & \textbf{0.067} \\\hline
\end{tabular}
\label{Table:Ablation_results}
\end{table}

\begin{figure}[h]
    \centering
    \includegraphics[width=\linewidth]{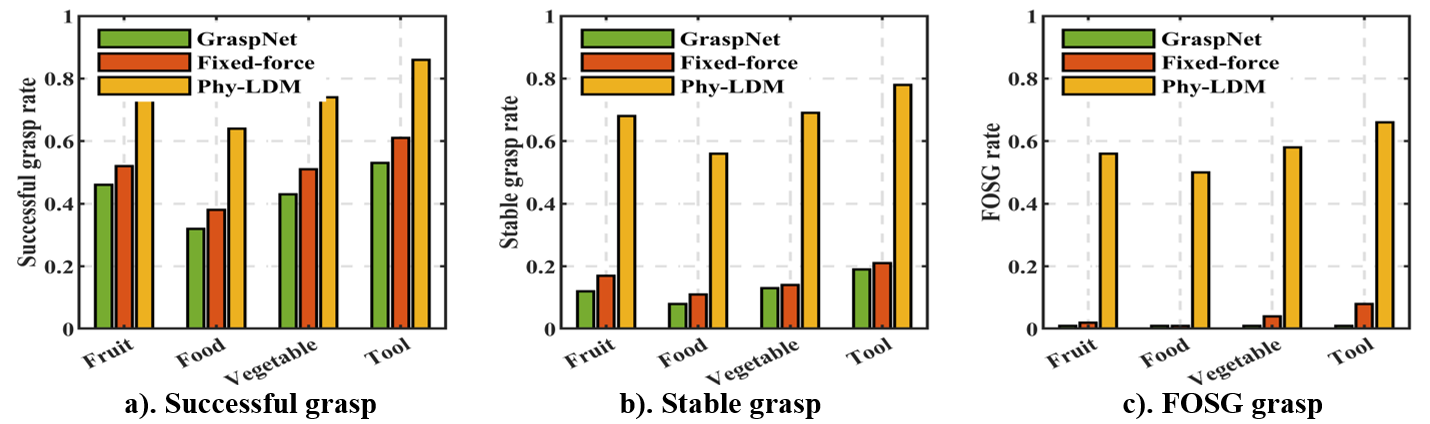}
    \caption{The experimental results for three grasp methods in four types of objects. Three evaluating metrics are calculated, i.e., successful grasp rate, stable grasp rate, and FOSG rate.}
    \label{fig:State}
\end{figure}

\subsection{Grasp experiment}
To demonstrate the effectiveness of our Phy-Tac, we implement our strategy on two robotic arms and compared it with  other classical methods, i.e., GraspNet and fixed-force method.
Specifically, the contact pose of our and fixed-force method are generated by the GraspNet model.
Furthermore, we consider three grasping states, i.e., successful grasp (SuG) which requites to catch the object, stable grasp (StG) which means successful grasp without slip and rotation, and FOSG which ensures optimal contact force in stable grasp.

\begin{figure}[h]
    \centering
    \includegraphics[width=\linewidth]{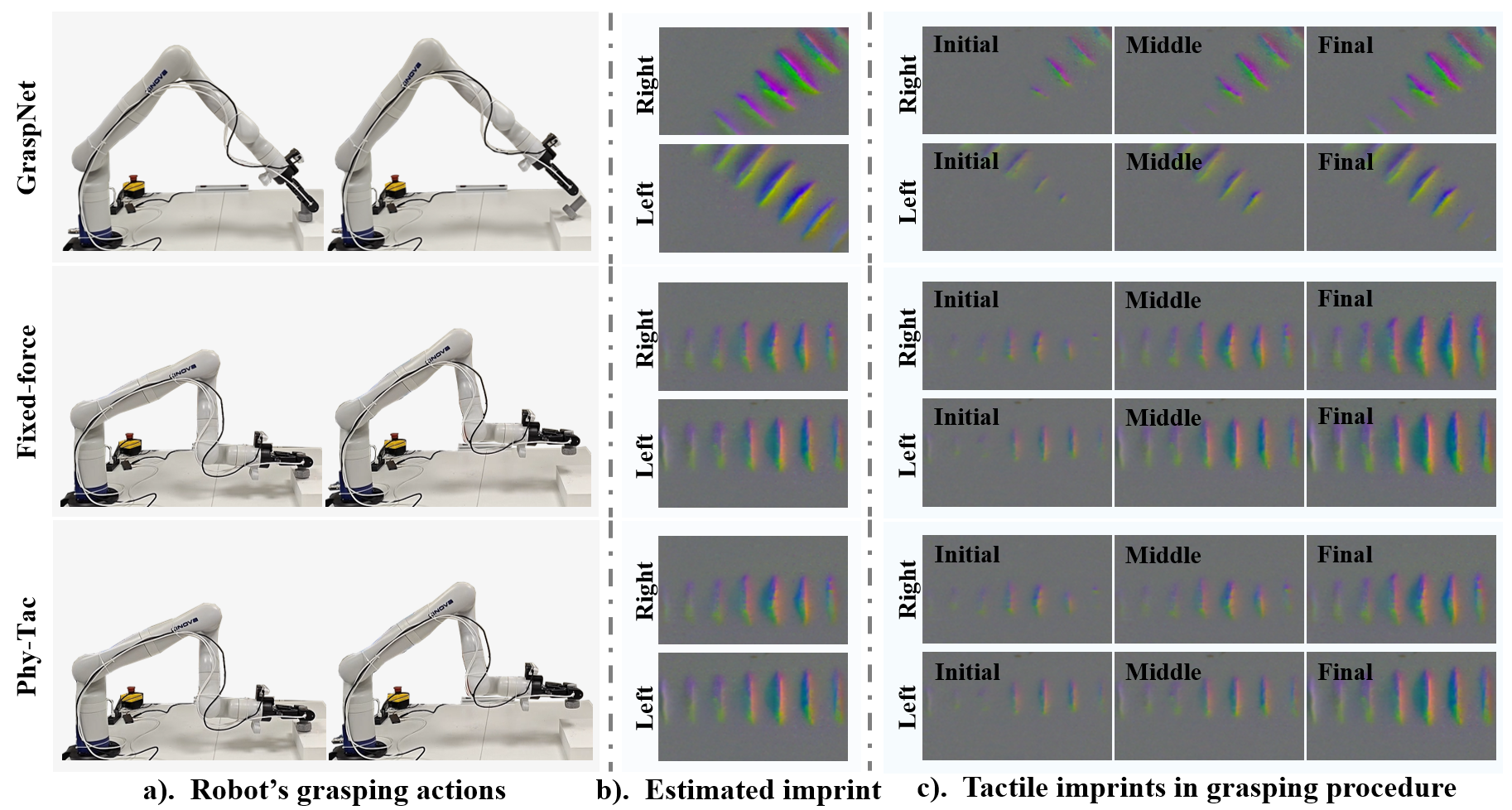}
    \caption{A robotic experiment for three grasping strategies. a). the initial contact and lifting states. b). the estimated imprint by Phy-LDM in initial contact condition. c). the tactile imprints during the grasping process.}
    \label{fig:experiments}
\end{figure}

As shown in Fig.~\ref{fig:State}, the successful rates of the three states over 50 grasp attempts indicate that the widely used fixed-force strategy can significantly improve the SuG rate ($\ge7\%$).
However, its effectiveness in enhancing states StG and FOSG is limited ($\le1\%$).
The reason to cause this results is that the given constant force can ensure that contact between gripper and object is real, which is not promised by original GraspNet.
Because the applied force is constant, this strategy cannot ensure the pre-setting is suitable for stable grasp
Fortunately, our method Phy-Tac can solve this shortcoming by it unique design, i.e., taking the force state (tactile image) into consideration.
A real-world experiment show in Fig.~\ref{fig:experiments} clearly demonstrates this: under GraspNet, the gripper makes contact but slips due to insufficient force; the fixed-force method achieves a grasp but with excessive deviation from the optimal force. In contrast, our Phy-Tac drives the gripper to apply the optimal contact force, successfully achieving FOSG.

Considering the challenges of tactile modeling for object with different material properties, especially the  deformable items, our Phy-Tac method demonstrates consistent and robust performance.
Among the four types of grasped objects, the three evaluation metrics of Phy-Tac show only slight variations as shown in Fig.~\ref{fig:State}.
For example, the differences between categories such as food (soft bread) and tools (hard screws) are not significant.
This is mainly because our method actively adjusts the gripper configuration to achieve optimized contact forces, thereby reducing the likelihood of deformation during stable grasping.
Consequently, Phy-LDM attains high success rates across objects with varying material properties.

\begin{figure}
    \centering
    \includegraphics[width=\linewidth]{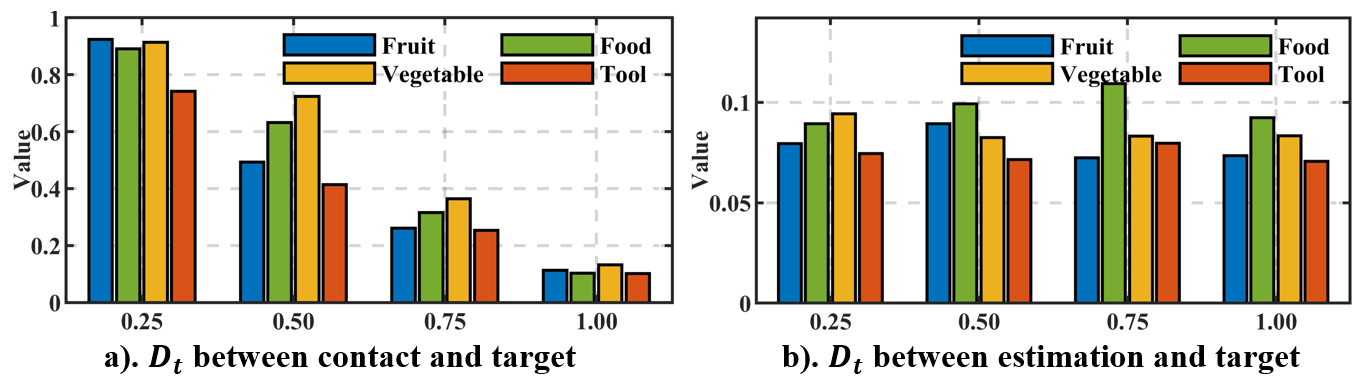}
    \caption{The error distance $D_t$ during the grasp process. The x-axis represents the percentage of the grasping process, and the y-axis represents the error distance $D_t$.
 }
    \label{fig:Results_tactile}
\end{figure}

To evaluate the LQR grasp servo, we visualize the error distance $D_t$ of one successful grasp for each type of object.
Specifically, the error distance $D_t$ between current tactile state during the grasp process and optimal tactile state under given grasp condition is shown in Fig.~\ref{fig:Results_tactile}-a.
In practice, we set the threshold $\delta_{in}$ as $0.15$.
The results indicate that our grasp servo can efficiently drive the robotic gripper to suitable position for achieving optimal tactile state, which requires $D_t\le\delta_{in}$. 
Therefore, our method is robust for all four types of objects.
In Fig.~\ref{fig:Results_tactile}-b, we indicate the $D_t$ for FOSG between the estimated optimal tactile in grasp process and the optimal tactile state under given grasp condition.
The small fluctuations of $D_t$ during the grasp process indicates that our Phy-LDM can estimated the optimal tactile for FOSG robustly, which benefits from incorporating object and contact-region physical information.
In summary, by leveraging physically informed tactile representations and an LQR-based servo controller, our method consistently converges to the optimal tactile state with minimal error, highlighting the critical role of physical priors in achieving robust and smooth grasp control.

\section{Conclusion}
\label{sec:Conclusion}
This paper presented a human-like framework, Phy-Tac, for FOSG that unifies pose planning, tactile prediction, and latent-space force control. 
A physics-conditioned latent diffusion model predicts the optimal tactile imprint, while an LQR-based controller drives the gripper toward force-optimal stability. 
Experiments on diverse rigid and compliant objects demonstrate that the proposed method significantly improves grasp stability and force efficiency compared with baseline strategies. 
Overall, this work establishes a pathway from stability-driven to force-optimal manipulation, contributing to safer and more adaptive tactile intelligence in robotic hands.

 \bibliographystyle{IEEEtran}
 \bibliography{IEEEabrv,References.bib}

\end{document}